\documentclass[letterpaper, 10 pt, conference]{ieeeconf}  

\IEEEoverridecommandlockouts                              

\usepackage{epsfig} 
\usepackage{times} 
\usepackage{amsmath} 
\usepackage{amssymb}  

\usepackage{graphicx}
\usepackage{bm}
\usepackage{blindtext}
\usepackage{caption}
\usepackage{microtype}

\newcommand{\ssto}[1][3pt]{\mathrel{%
		\hbox{\rule[\dimexpr\fontdimen22\textfont2-.2pt\relax]{#1}{.4pt}}%
		\mkern-4mu\hbox{\usefont{U}{lasy}{m}{n}\symbol{41}}}}

\title{\LARGE \bf
Learning Feature Decomposition for Domain Adaptive Monocular Depth Estimation
}

\author{Shao-Yuan Lo$^{1}$, Wei Wang$^{2}$, Jim Thomas$^{2}$, Jingjing Zheng$^{2}$, Vishal M. Patel$^{1}$, and Cheng-Hao Kuo$^{2}$  
\thanks{$^{1}$ Shao-Yuan Lo and Vishal M. Patel are with the Department of Electrical and Computer Engineering, Johns Hopkins University, Baltimore, MD, USA. {\tt\small \{sylo, vpatel36\}@jhu.edu}}
\thanks{$^{2}$ Wei Wang, Jim Thomas, Jingjing Zheng, and Cheng-Hao Kuo are with Amazon Lab126. {\tt\small \{wweiwan, jimthoma, zhejingj, chkuo\}@amazon.com}}
\thanks{* This work was mostly done when Shao-Yuan Lo was an intern at Amazon Lab126.}
}

\begin{document}

\maketitle
\thispagestyle{empty}
\pagestyle{empty}

\begin{abstract}
Monocular depth estimation (MDE) has attracted intense study due to its low cost and critical functions for robotic tasks such as localization, mapping and obstacle detection. Supervised approaches have led to great success with the advance of deep learning, but they rely on large quantities of ground-truth depth annotations that are expensive to acquire. Unsupervised domain adaptation (UDA) transfers knowledge from labeled source data to unlabeled target data, so as to relax the constraint of supervised learning. However, existing UDA approaches may not completely align the domain gap across different datasets because of the domain shift problem. We believe better domain alignment can be achieved via well-designed feature decomposition. In this paper, we propose a novel UDA method for MDE, referred to as Learning Feature Decomposition for Adaptation (LFDA), which learns to decompose the feature space into content and style components. LFDA only attempts to align the content component since it has a smaller domain gap. Meanwhile, it excludes the style component which is specific to the source domain from training the primary task. Furthermore, LFDA uses separate feature distribution estimations to further bridge the domain gap. Extensive experiments on three domain adaptative MDE scenarios show that the proposed method achieves superior accuracy and lower computational cost compared to the state-of-the-art approaches.
\end{abstract}

\section{INTRODUCTION} \label{sec:intro}
Depth information is essential to many robotic applications, \textit{e.g.}, localization, mapping and obstacle detection. Existing depth acquisition devices, such as Lidar and structured-light sensors, are typically bulky, heavy and power-consuming. Therefore, they are unsuitable for compact robotic platforms. This motivates the progress of monocular depth estimation (MDE) that predicts depth from a single image, as it has low cost, small size, high power efficiency, and is no need to re-calibrate after a long time of use.

Recent advances in deep learning have enabled supervised learning approaches to perform MDE \cite{bhat2021adabins, eigen2014depth, fu2018deep, laina2016deeper}, but obtaining ground-truth depth annotations are costly and labor-intensive. Moreover, the obtained depth annotations only correspond to that specific camera. In other words, a model trained with the images and annotations of a specific camera may not generalize well to another camera with different camera settings, \textit{e.g.}, focal length and size of field view. Synthetic data and their annotations are easier to acquire. However, the supervised approaches trained on such a synthetic dataset often suffer from severe accuracy degradation when tested on real data, as different datasets have distinct characteristics. This is known as the domain shift problem. These challenges hinder the MDE technique from applying to compact robotic platforms. Hence, developing algorithms that can transfer the knowledge learned from one labeled dataset to another unlabeled dataset becomes increasingly important.

We approach this via unsupervised domain adaptation (UDA) in which given a labeled source dataset and an unlabeled target dataset, the objective is to learn a MDE model that has a good generalization to the unlabeled target domain data. Existing works mainly rely on a synthetic-to-real translation or vice versa to bridge the domain gap \cite{akada2021self,atapour2018real,chen2019crdoco,lopez2020desc,pnvr2020sharingan,zhao2019geometry,zheng2018t2net}. Although these works have achieved great improvements, image translation itself is not an easy task. Images may not be perfectly translated to another domain or contain distortion after translation. Another research stream performs feature alignment through adversarial learning \cite{chen2019crdoco,kundu2018adadepth,pnvr2020sharingan,zheng2018t2net}. Nevertheless, it is difficult to completely align the entire feature space from different domains owing to the domain shift problem.

\begin{figure}
	\centering
	\includegraphics[width=0.48\textwidth]{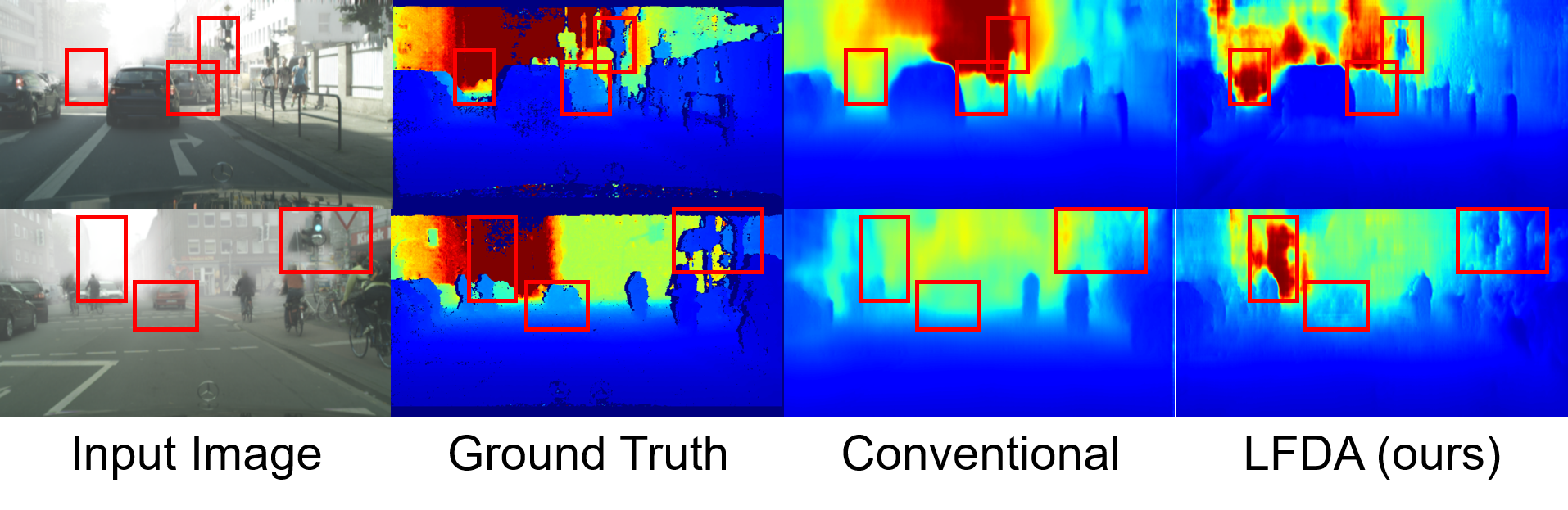}
	\caption{Example results of domain adaptive MDE on the Foggy Cityscapes dataset \cite{sakaridis2018semantic}. It is a scenario of adverse weather adaptation. ``Conventional" refers to the method based on the usual domain adversarial learning \cite{ganin2015unsupervised}. The red boxes highlight regions where our method makes improvements.}
	\label{fig:cover}
\end{figure}

To overcome these challenges, inspired by recent approaches \cite{pnvr2020sharingan,zheng2018t2net} and disentangled learning techniques \cite{chang2019all,lai20eccv,lee20icra,li2019cross,liu18cvpr}, we assume that the feature space can be decomposed into content and style components. The content component consists of semantic features that are shared across different domains. For example, consider images of indoor scenes from two different datasets. Objects like tables, chairs and beds are content information. Such semantic features are more domain-invariant, so it is easier to align the content component from different domains. In contrast, the style component is domain-specific. For instance, style features like texture and color are unique to the scenes captured by a particular camera, so aligning the style features may not be practical. Hence, to train a MDE model working for the target data, we suggest to discard the source-specific style component that hinders adaptation to narrow the domain gap, but include the target-specific style component that is still useful for the primary MDE task.

Based on the above intuitions, we propose a novel UDA method for the MDE task, referred to as Learning Feature Decomposition for depth Adaptation (LFDA): (1) Different from prior works attempting to align the entire feature maps of source and target data \cite{chen2019crdoco,kundu2018adadepth,pnvr2020sharingan,zheng2018t2net}, LFDA only needs to align the content features that already have a much smaller domain gap. (2) To further improve the content feature alignment, LFDA individually estimates the statistics of different feature domains via separate batch normalizations (BNs) \cite{chang2019domain,lo2021defending,xie2020adversarial}, which can bypass the domain-specific elements in the feature space. The separate BN structure also helps to properly integrate the content and style features of the target data. (3) With the proposed decomposition learning, LFDA bridges the domain gap more efficiently. In particular, it keeps a relatively compact structure at inference time, leading to lower computational complexity compared to the recent advances which require a sophisticated image translation network during inference \cite{pnvr2020sharingan,zhao2019geometry}. (4) In addition, most existing approaches rely on a multi-stage training procedure that first pre-trains each sub-networks separately then fine-tunes them together \cite{akada2021self,atapour2018real,lopez2020desc,pnvr2020sharingan,zhao2019geometry}. Instead, LFDA is trained end-to-end in a single stage, making it more feasible to deploy in practical applications. 

In evaluation, the majority of existing studies only focus on synthetic-to-real adaptation \cite{akada2021self,atapour2018real,chen2019crdoco,kundu2018adadepth,pnvr2020sharingan,zhao2019geometry,zheng2018t2net}. In contrast, we apply our method to three broad scenarios of domain adaptation: (1) cross-camera adaptation, (2) synthetic-to-real adaptation, and (3) adverse weather adaptation \cite{vs2021mega}. To the best of our knowledge, this paper is the first attempt that considers all the three scenarios for the MDE task. Particularly, adverse weather adaptation is the first time explored for MDE. Fig.~\ref{fig:cover} shows examples of adverse weather adaptation results. Compared to a conventional approach, our LFDA can obtain more accurate depth predictions for cars, traffic signs, sky, \textit{etc.}, under foggy weather condition. More extensive experiments in Sec.~\ref{sec:expt} demonstrate that LFDA achieves promising performance on all the scenarios.

\section{RELATED WORK}
\noindent \textbf{Monocular depth estimation (MDE).}
Deep learning has achieved high accuracy for MDE by supervised learning. Eigen \textit{et al.} \cite{eigen2014depth} introduced a deep learning-based MDE approach with a multi-scale network. Afterward, Laina \textit{et al.} \cite{laina2016deeper} presented a deeper network with a fully convolutional network and residual learning. Fu \textit{et al.} \cite{fu2018deep} divided depth ranges into multiple depth bins and solved MDE in a classification manner using an ordinal regression loss. Recently, Bhat \textit{et al.} \cite{bhat2021adabins} developed a transformer-based block to adaptively adjust the depth bins for each image. Several studies explore training MDE models via self-supervision. Notable algorithms include exploiting epipolar geometry constraints from stereo pairs \cite{garg2016unsupervised,godard2017unsupervised,kuznietsov2017semi} and utilizing multi-view information from monocular video sequences \cite{mahjourian2018unsupervised,zhou2017unsupervised}.

\noindent \textbf{Unsupervised domain adaptation (UDA).}
Because depth annotations are prohibitively expensive for supervised learning, unsupervised domain adaptation has gained a lot of interest in the research community. Recent approaches involve feature distribution alignment using adversarial learning \cite{ganin2015unsupervised,tzeng2017adversarial}, image-to-image translation \cite{hoffman2018cycada,murez2018image}, and pseudo-label generation \cite{chen2019progressive,saito2017asymmetric}.

\noindent \textbf{Domain adaptive MDE.}
Domain adaptation for MDE was introduced by Atapour \textit{et al.} \cite{atapour2018real}, where they trained a depth estimation network using synthetic images then translated real images to synthetic style during inference. AdaDepth \cite{kundu2018adadepth} employs adversarial learning at both feature and output spaces to align the distributions between the source and target domains. T$^2$Net \cite{zheng2018t2net} transfers synthetic images to real style to train a MDE network. CrDoCo \cite{chen2019crdoco} and GASDA \cite{zhao2019geometry} use bidirectional style transfer to learn the mapping between two domains, where GASDA also exploits epipolar geometry structure for real images. SharinGAN \cite{pnvr2020sharingan} translates both synthetic and real data to a single shared domain to decrease their discrepancy. DESC \cite{lopez2020desc} leverages an additional semantic segmentation network and edge detection to provide semantic and edge guidance. Akada et al. \cite{akada2021self} adopted recent SSRL techniques to learn domain-invariant representations. However, they either suffer from sub-optimal domain alignment or high computational complexity during inference.

\begin{figure*}
	\centering
	\includegraphics[width=0.96\textwidth]{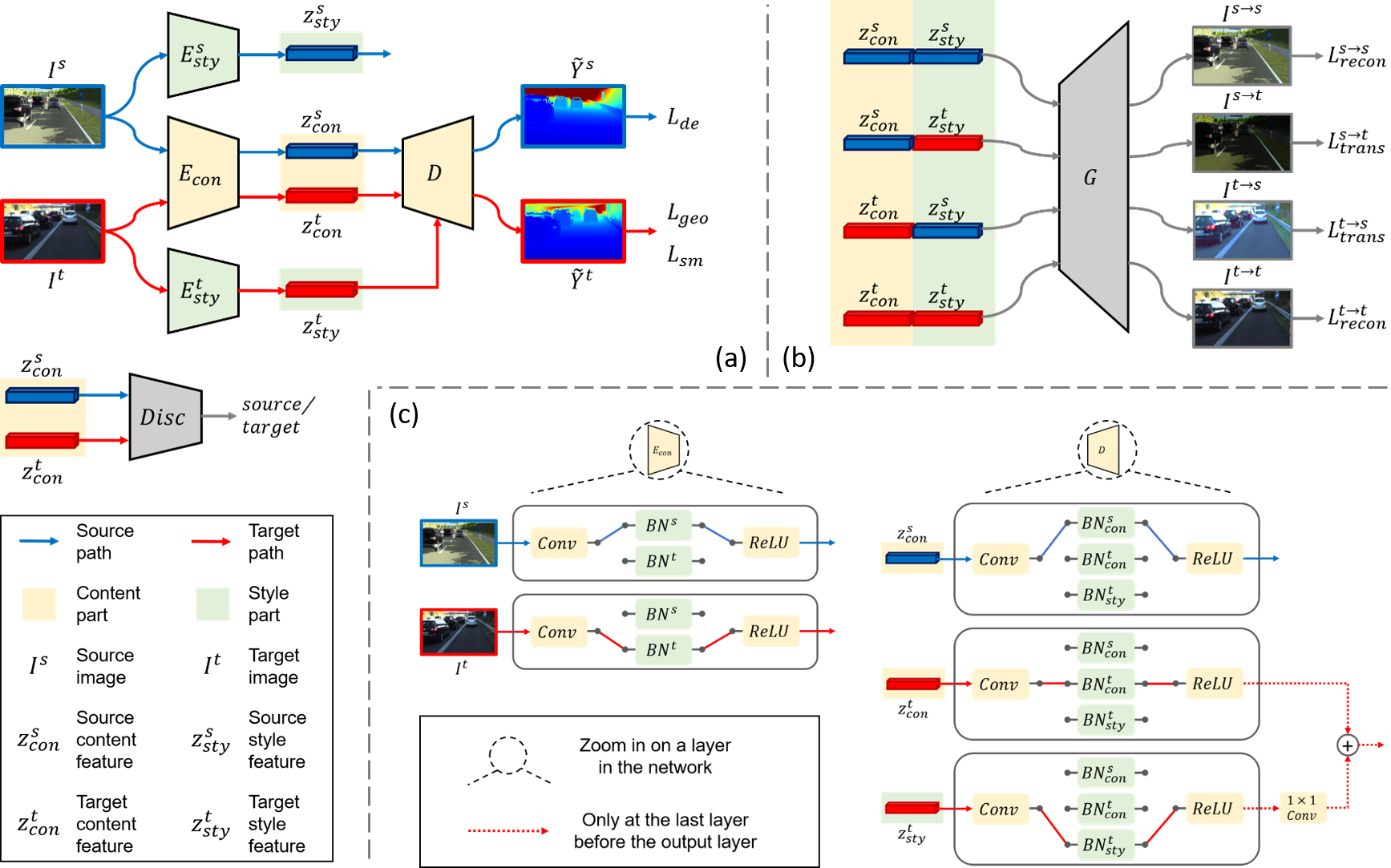}
	\caption{Overview of the proposed LFDA framework. $E_{con}$: shared content encoder, $E^s_{sty}$: source-specific style encoder, $E^t_{sty}$: target-specific style encoder, $D$: depth estimation task decoder, $G$: generator, and $Disc$: domain discriminator. (a) Main information flow. (b) Learning translations for feature decomposition. (c) Separate BN structure for feature alignment and integration.}
	\label{fig:big_pic}
\end{figure*}

\section{PROPOSED METHOD}

\subsection{Framework}
An overview of the proposed LFDA is shown in Fig.~\ref{fig:big_pic}. The entire framework consists of eight sub-networks: shared content encoder $E_{con}$, source-specific style encoder $E^s_{sty}$, target-specific style encoder $E^t_{sty}$, MDE task decoder $D$, generator $G$, domain discriminator $Disc$, source-to-target translation discriminator $Disc^{s \ssto t}$, and target-to-source translation discriminator $Disc^{t \ssto s}$. $\{E_{con}, D\}$ composes as a MDE primary task network, which is a standard encoder-decoder architecture.

\noindent \textbf{Feature decomposition.}
As illustrated in Fig.~\ref{fig:big_pic}a, the two individual style encoders $E^s_{sty}$ and $E^t_{sty}$ extract the domain-specific style features of the given source input $I^s$ and target input $I^t$, respectively. This is formulated as $z^s_{sty} = E^s_{sty}(I^s)$ and $z^t_{sty} = E^t_{sty}(I^t)$. We believe that the content of images is more domain-invariant, so a shared content encoder $E_{con}$ is used to learn the content features of both source and target images, formulated as $z^s_{con} = E_{con}(I^s)$ and $z^t_{con} = E_{con}(I^t)$. This decomposition is achieved by the training scheme shown in Fig.~\ref{fig:big_pic}b, and the details are elaborated in Sec.~\ref{sec:objective}.

\noindent \textbf{Feature alignment.}
Although the content features $z^s_{con}$ and $z^t_{con}$ learned by a standard encoder already have a small domain gap, they are still not completely domain-invariant, as the content of images from different domains also contains some domain-specific elements, such as scale and viewpoint. To address this, we perform feature alignment in two aspects.

First, we propose to estimate the feature distributions of $I^s$ and $I^t$ individually using a separate BN structure \cite{chang2019domain,lo2021defending,xie2020adversarial}. Specifically, two BN branches \cite{ioffe2015batch}, denoted as $BN^s$ and $BN^t$, are deployed after each convolutional layer in $E_{con}$ (see Fig.~\ref{fig:big_pic}c). Each BN branch works individually for its own domain. To elaborate, $BN^s$ and $BN^t$ learn domain-specific affine parameters $\{\gamma^s, \beta^s\}$/$\{\gamma^t, \beta^t\}$, and distribution statistics $\{\mu^s, \sigma^s\}$/$\{\mu^t, \sigma^t\}$ for the source and target data, respectively. Note that all the layers other than BNs are still shared (\textit{e.g.}, convolution and ReLU). Suppose that $\ddot{z}^d_{con}$ is the content feature of domain $d$, where $d \in \{s, t\}$, the separate BN structure at an arbitrary layer in $E_{con}$ is formulated as:
\begin{equation}
\label{sbn_src}
BN^d(\ddot{z}^d_{con}; \, \gamma^d, \beta^d) = \gamma^d \, \bigg( \frac{\ddot{z}^d_{con} - \mu^d}{\sqrt{(\sigma^d)^2 + k}} \bigg) + \beta^d,
\end{equation}
where $k$ is a tiny constant for numerical stability. With this design, the domain gap between $z^s_{con}$ and $z^t_{con}$ is acquired by the domain-specific parameters $\{\mu ^d, \sigma^d, \gamma^d, \beta^d\}$, and their domain-invariant part passes through each BN layer.

Second, inspired by GRL \cite{ganin2015unsupervised}, we employ adversarial learning \cite{goodfellow2014generative} to align the features $z^s_{con}$ and $z^t_{con}$ (see Fig.~\ref{fig:big_pic}a). Details are discussed in Sec.~\ref{sec:objective}.

\noindent \textbf{Feature integration.}
Our feature decomposition extracts four preferred components: $\{z^s_{con}, z^s_{sty}, z^t_{con}, z^t_{sty}\}$, where $z^s_{con}$ and $z^t_{con}$ are aligned by our separate BN structure and adversarial learning. To train the MDE task decoder $D$, we use $z^s_{con}$, $z^t_{con}$ and $z^t_{sty}$. We discard $z^s_{sty}$ since it is specific to source data and thus cannot help the model adapt to the target domain. Instead, the target-specific style component $z^t_{sty}$ is still useful for the MDE model that works for the target domain.

After feature decomposition, $z^t_{sty}$ and $z^t_{con}$ have different feature characteristics though they are from the same target domain. Hence, directly fusing them in the task decoder $D$ would cause potential accuracy degradation. To address this issue, as shown in Fig.~\ref{fig:big_pic}c, we also deploy separate BNs in $D$. There are three BN branches: $BN^s_{con}$, $BN^t_{con}$ and $BN^t_{sty}$, each of which works as Eq.~(\ref{sbn_src}). $BN^s_{con}$ and $BN^t_{con}$ are used for the same purpose as discussed before, and $BN^t_{sty}$ is responsible for characterizing the feature distribution of $z^t_{sty}$ exclusively. Since the content and style features have different underlying distributions, simply leveraging a single set of BN parameters for $z^t_{con}$ and $z^t_{sty}$ would estimate an inaccurate mixture. Therefore, the additional $BN^t_{sty}$ is used to disentangle such mixture distribution, allowing proper integration of $z^t_{con}$ and $z^t_{sty}$ for decoding target features. Because the content and style components may have different importance for MDE, we employ a $1 \times 1$ convolution and a residual connection to combine $z^t_{con}$ and $z^t_{sty}$ right before the output layer of $D$. This weighted fusion helps to adjust the balance between these two features of target data. Finally, $D$ outputs predicted depth maps, $\tilde{Y}^s=D(z^s_{con})$ and $\tilde{Y}^t=D(z^t_{con}, z^t_{sty})$, respectively.

\subsection{Objectives} \label{sec:objective}
The proposed LFDA framework is trained with the following objective functions.

\noindent \textbf{Feature decomposition loss.}
This loss is used to decompose the feature components according to our assumption for domain adaptation. It consists of translation loss and reconstruction loss.

Inspired by style transfer techniques \cite{huang2017arbitrary,johnson2016perceptual}, we adopt the translation loss to separate the content and style features of an input images. Let us consider the case of source-to-target image translation in our framework. Given a source image $I^s$ and a target image $I^t$, we aim to derive a translated image $I^{s \ssto t} = G(z^s_{con}, z^t_{sty})$ which consists of the content of $I^s$ and the style of $I^t$ (see Fig.~\ref{fig:big_pic}b). We achieve this translation via objective $\mathcal{L}^{s \ssto t}_{trans}$, which consists of two perceptual losses \cite{johnson2016perceptual} and an adversarial loss:
\begin{equation}
\label{trans_loss_src}
\begin{split}
\mathcal{L}^{s \ssto t}_{trans}
& = \sum_{j \in L} w^{trans}_{con, j} \big\| \phi_j(I^s) - \phi_j(I^{s \ssto t}) \big\|_1 \\
& + \sum_{j \in L} w^{trans}_{sty, j} \big\| \mu(\phi_j(I^t)) - \mu(\phi_j(I^{s \ssto t})) \big\|_1 \\
& + \eta \, \big(Disc^{s \ssto t}(I^{s \ssto t}) - 1 \big)^2,
\end{split}
\end{equation}
where $\eta = 0.2$, $w^{trans}_{con}$ and $w^{trans}_{sty}$ are pre-defined weights, $L$ denotes the $\{relu1\_1, relu2\_1, relu3\_1, relu4\_1, relu5\_1\}$ layers of a pre-trained VGG network \cite{simonyan2015very} that measures perceptual loss, $\phi_j$ is the $j$-th layer in $L$, and $\mu(\cdot)$ returns the channel-wise mean values of a feature space. This translation loss has also been explored by \cite{chang2019all}.

To elaborate, the first perceptual loss computes the distance of the high-level content features between $I^s$ and $I^{s \ssto t}$ such that $I^{s \ssto t}$ contains the content of $I^s$. Since the content information mostly exists in higher layers of VGG, we set $w^{trans}_{con}$ to $\{0, 0, 0, 1/4, 1\}$. The second perceptual loss forces $I^{s \ssto t}$ to contain the style of $I^t$. To explicitly encode the style information of an image, we employ AdaIN structure \cite{huang2017arbitrary} that measures the distance of the channel-wise mean values of the style features between $I^t$ and $I^{s \ssto t}$. Since the style information mostly exists in lower layers of VGG, we set $w^{trans}_{sty}$ to $\{1, 1, 1, 0, 0\}$. The third term is a standard least-squares adversarial loss \cite{mao2017least}, where we assign labels 1 and 0 to untranslated and translated images, respectively. This loss helps to improve the quality of image translation. As for the case of target-to-source translation, it is symmetric to source-to-target translation. We define its objective as $\mathcal{L}^{t \ssto s}_{trans}$, which replaces $s$ to $t$, $t$ to $s$ and $s \ssto t$ to $t \ssto s$ in Eq.~(\ref{trans_loss_src}).

The reconstruction loss is used to guarantee that the combination of the decomposed content and style components forms a nearly complete representation of an input image \cite{chang2019all}. Let us consider the case of source image reconstruction. Given a source image $I^s$, we aim to derive a reconstruction $I^{s \ssto s} = G(z^s_{con}, z^s_{sty})$ (see Fig.~\ref{fig:big_pic}b). This can be achieved via objective $\mathcal{L}^{s \ssto s}_{recon}$, which is also based on the perceptual loss:
\begin{equation}
\label{recon_loss_src}
\mathcal{L}^{s \ssto s}_{recon}
= \sum_{j \in L} w^{recon}_j \big\| \phi_j(I^s) - \phi_j(I^{s \ssto s}) \big\|_1,
\end{equation}
where $w^{recon} = \{1/32, 1/16, 1/8, 1/4, 1\}$. Symmetrically, target image reconstruction is achieved via objective $\mathcal{L}^{t \ssto t}_{recon}$, which replaces $s$ to $t$ and $s \ssto s$ to $t \ssto t$, from Eq.~(\ref{recon_loss_src}).

With the above loss functions, LFDA decomposes the feature space into $\{z^s_{con}, z^s_{sty}, z^t_{con}, z^t_{sty}\}$, where each of which contains its supposed information exclusively.

\noindent \textbf{Feature alignment loss.}
Different from prior works that attempt to align the entire features \cite{chen2019crdoco,kundu2018adadepth,pnvr2020sharingan,zheng2018t2net}, LFDA only needs to align the content features that already have a much smaller domain gap, which is easier to achieve.
Inspired by GRL \cite{ganin2015unsupervised}, we use a domain adversarial loss $\mathcal{L}_{align}$ to align the distributions of $z^s_{con}$ and $z^t_{con}$ (see Fig.~\ref{fig:big_pic}a). This is defined as: $\mathcal{L}_{align} = \big( Disc(z^s_{con}) \big)^2 + \big( Disc(z^t_{con}) - 1 \big)^2$, where we assign labels 1 and 0 to the source and target domain, respectively. We use the least-squares adversarial loss \cite{mao2017least} because it is shown to be more stable at training time. Eventually, $\mathcal{L}_{align}$ further reduces the discrepancy between $z^s_{con}$ and $z^t_{con}$.

\noindent \textbf{Depth estimation loss.}
This is the primary task objective for MDE. We employ $L_1$ loss to make use of the source data annotations: $\mathcal{L}^s_{de}= \| \tilde{Y}^s - Y^s \|_1$, where $\tilde{Y}^s = D(z^s_{con})$ is the predicted depth map and $Y^s$ is the corresponding ground-truth. Following GASDA \cite{zhao2019geometry} and SharinGAN \cite{pnvr2020sharingan}, depth smoothness loss $\mathcal{L}_{sm}$ and geometry consistency loss $\mathcal{L}_{geo}$ are used as self-supervisions for the target data. They are defined as: $\mathcal{L}_{sm} = e^{-\nabla I^t} \| \nabla \tilde{Y}^t \|_1$, where $\tilde{Y}^t = D(z^t_{con}, z^t_{sty})$ is the predicted depth map; $\mathcal{L}_{geo} = \alpha \big( 1 - SSIM(I^t, \hat{I}^t) \big) + \beta \| I^t - \hat{I}^t \|_1$, where $\alpha = 0.425$, $\beta = 0.15$, $\hat{I}^t$ is the inverse warped image derived from $\tilde{Y}^t$ the right counterpart of $I^t$, and SSIM \cite{wang2004image} is an image quality metric. Moreover, inspired by image translation-based adaptation approaches \cite{akada2021self,chen2019crdoco,lopez2020desc,zhao2019geometry,zheng2018t2net}, we leverage $I^{s \ssto t}$ that is generated during feature decomposition learning, to adapt the task network to the target domain (\textit{i.e.}, feed $I^{s \ssto t}$ produced from Fig.~\ref{fig:big_pic}b into the pipeline of Fig.~\ref{fig:big_pic}a). This is defined as: $z^{s \ssto t}_{con} = E_{con}(I^{s \ssto t})$, $z^{s \ssto t}_{sty} = E^t_{sty}(I^{s \ssto t})$, and $\hat{Y}^{s \ssto t} = D(z^{s \ssto t}_{con}, z^{s \ssto t}_{sty})$. Then, the $L_1$ loss is used to train with the translated image: $\mathcal{L}^{s \ssto t}_{de} = \| \tilde{Y}^{s \ssto t} - Y^s \|_1$.

\noindent \textbf{Full learning objective.}
The full objective of the proposed LFDA framework is defined as:
\begin{equation}
\label{full_loss}
\begin{split}
\mathcal{L}
& = (\mathcal{L}^s_{de} + \mathcal{L}^{s \ssto t}_{de}) + \lambda_{geo} \mathcal{L}_{geo} + \lambda_{sm} \mathcal{L}_{sm} + \lambda_{align} \mathcal{L}_{align} \\
& + \lambda_{recon} (\mathcal{L}^{s \ssto s}_{recon} + \mathcal{L}^{t \ssto t}_{recon}) + \lambda_{trans} (\mathcal{L}^{s \ssto t}_{trans} + \mathcal{L}^{t \ssto s}_{trans}),
\end{split}
\end{equation}
where $\lambda$'s are trade-off factors. We optimize this loss function end-to-end in a single stage.

\subsection{Inference}
During inference, our goal is to predict a depth map from a given target image. This corresponds to the red path in Fig.~\ref{fig:big_pic}a. Therefore, only $E_{con}$, $E^t_{sty}$ and $D$ are retained after training, where $E^t_{sty}$ is the only required sub-network in addition to the MDE primary task network $\{E_{con}, D\}$. Compared to recent top-performing approaches which require an entire sophisticated image translation network during inference \cite{pnvr2020sharingan,zhao2019geometry}, LFDA allows much lower computational complexity. This is attributed to the proposed decomposition learning that reduces the domain gap more efficiently.

\section{EXPERIMENTS} \label{sec:expt}
We extensively evaluate the proposed LFDA on three domain adaptation scenarios: cross-camera adaptation, synthetic-to-real adaptation, and adverse weather adaptation \cite{vs2021mega}. Moreover, we conduct an ablation study and analyze the computational complexity of the models.

\subsection{Implementation Details}
For fair comparison, the architectures of sub-networks $\{E_{con}, D\}$, $E^s_{sty}$, $E^t_{sty}$, $Disc$, $Disc^{t \ssto s}$ and $Disc^{s \ssto t}$ are implemented identical to the corresponding ones in T$^2$Net \cite{zheng2018t2net}. Besides, generator $G$ is implemented as in \cite{chang2019all}. The models are trained by Adam optimizer \cite{kingma2015adam} with initial learning rates of $1e^{-4}$ for $\{E_{con}, D\}$ and $2e^{-5}$ for the other sub-networks. The learning rates decrease according to the polynomial decay policy. We set $\lambda_{geo}=1$, $\lambda_{sm}=\lambda_{align}=0.01$, $\lambda_{recon}=0.5$, and $\lambda_{trans}=0.05$. The entire framework is trained end-to-end in a single stage. The experiments are implemented by PyTorch \cite{paszke2019pytorch} and conducted on a single NVIDIA Tesla V100 GPU. We will release our source code after the paper gets accepted.

\renewcommand{\arraystretch}{0.9}
\captionsetup{skip=0pt}
\setlength{\tabcolsep}{4.5pt}
\begin{table}[t!]
	\scriptsize
	\begin{center}
		\caption{Results of Cityscapes-to-KITTI adaptation, tested on KITTI Eigen split \cite{eigen2014depth} (cap 80m). The 1.25$^n$ columns refer to the standard $\delta <$ 1.25$^n$ accuracy metrics.}
		\label{table:city2kitti}
		\begin{tabular}{l | cccc | ccc}
			\hline \noalign{\smallskip} \noalign{\smallskip}
			&  & \multicolumn{2}{c}{\underline{\hspace{0.4em} Lower, better \hspace{0.4em}}} &  & \multicolumn{3}{c}{\underline{\hspace{0.4em} Higher, better \hspace{0.4em}}} \\
			Method & abs-rel & sq-rel & rmse & rmse-log & 1.25 & 1.25$^2$ & 1.25$^3$ \\
			\noalign{\smallskip} \hline \noalign{\smallskip}
			T$^2$Net \cite{zheng2018t2net} & 0.173 & 1.335 & 5.640 & 0.242 & 0.773 & 0.930 & 0.970 \\
			DESC \cite{lopez2020desc} & 0.149 & 0.967 & 5.236 & 0.223 & 0.810 & 0.940 & 0.976 \\ 
			\noalign{\smallskip} \hline \noalign{\smallskip}
			LFDA (ours) & \textbf{0.119} & \textbf{0.963} & \textbf{5.049} & \textbf{0.207} & \textbf{0.855} & \textbf{0.948} & \textbf{0.977} \\
			\noalign{\smallskip} \hline
		\end{tabular}
	\end{center}
	\vspace*{-\baselineskip}
\end{table}

\subsection{Cross-Camera Adaptation}
Different cameras may have distinct intrinsic parameters or viewpoints, making the captured images have different scales, fields of view, \textit{etc}. Such domain gap could cause sub-optimal adaptation performance.

\noindent \textbf{Datasets.}
We use Cityscapes \cite{Cordts2016Cityscapes} as the source dataset and KITTI \cite{geiger2012we} as the target dataset. The KITTI Eigen split \cite{eigen2014depth} is used for testing. Following \cite{zheng2018t2net}, we rescale the input size of KITTI images from 375×1242 to 192×640, and upsample the predicted depth maps to the original size for evaluation. For Cityscapes, we follow \cite{lopez2020desc} that crops and resizes the images from 1024×2048 to 192×640. The ground-truth depth is capped at 80m.

\noindent \textbf{Results.}
Table~\ref{table:city2kitti} reports the results adhered to a standard evaluation protocol \cite{eigen2014depth}. The impressive improvements on all the metrics show the superiority of our LFDA. In particular, LFDA's abs-rel error is 20\% lower than DESC \cite{lopez2020desc}. This indicates that the proposed learning of feature decomposition is effective to reduce the domain gap between the images captured by different cameras.

\setlength{\tabcolsep}{2pt}
\begin{table}[t!]
	\scriptsize
	\begin{center}
		\caption{Results of X-to-KITTI adaptation, tested on KITTI stereo 2015 \cite{menze2015object}. Top-2 methods are in bold. vK: Virtual KITTI, K: KITTI, CS: Cityscapes, G: GTA5 images.}
		\label{table:kitti2015}
		\begin{tabular}{l | c | cccc | ccc}
			\hline \noalign{\smallskip} \noalign{\smallskip}
			&  &  & \multicolumn{2}{c}{\underline{\hspace{0.4em} Lower, better \hspace{0.4em}}} &  & \multicolumn{3}{c}{\underline{\hspace{0.4em} Higher, better \hspace{0.4em}}} \\
			Method & Dataset & abs-rel & sq-rel & rmse & rmse-log & 1.25 & 1.25$^2$ & 1.25$^3$ \\
			\noalign{\smallskip} \hline \noalign{\smallskip}
			Atapour \textit{et al.} \cite{atapour2018real} & G $\ssto$ K & 0.101 & 1.048 & 5.308 & 0.184 & 0.903 & \textbf{0.988} & \textbf{0.992} \\
			GASDA \cite{zhao2019geometry} & vK $\ssto$ K & 0.106 & \textbf{0.987} & 5.215 & 0.176 & 0.885 & 0.963 & \textbf{0.986} \\
			\noalign{\smallskip} \hline \noalign{\smallskip}
			LFDA (ours) & CS $\ssto$ K & \textbf{0.092} & 1.055 & \textbf{5.024} & \textbf{0.165} & \textbf{0.906} & 0.966 & 0.985 \\
			LFDA (ours) & vK $\ssto$ K & \textbf{0.087} & \textbf{0.931} & \textbf{4.765} & \textbf{0.162} & \textbf{0.910} & \textbf{0.968} & \textbf{0.986} \\
			\noalign{\smallskip} \hline
		\end{tabular}
	\end{center}
	\vspace*{-\baselineskip}
\end{table}

\setlength{\tabcolsep}{4pt}
\begin{table}[t!]
	\scriptsize
	\begin{center}
		\caption{Results of vKITTI-to-KITTI adaptation, tested on KITTI Eigen split \cite{eigen2014depth} (cap 80m). Top-2 methods are in bold.}
		\label{table:vkitti2kitti}
		\begin{tabular}{l | cccc | ccc}
			\hline \noalign{\smallskip} \noalign{\smallskip}
			&  & \multicolumn{2}{c}{\underline{\hspace{0.4em} Lower, better \hspace{0.4em}}} &  & \multicolumn{3}{c}{\underline{\hspace{0.4em} Higher, better \hspace{0.4em}}} \\
			Method & abs-rel & sq-rel & rmse & rmse-log & 1.25 & 1.25$^2$ & 1.25$^3$ \\
			\noalign{\smallskip} \hline \noalign{\smallskip}
			AdaDepth \cite{kundu2018adadepth} & 0.214 & 1.932 & 7.157 & 0.295 & 0.665 & 0.882 & 0.950 \\
			CrDoCo \cite{chen2019crdoco} & 0.232 & 2.204 & 6.733 & 0.291 & 0.739 & 0.883 & 0.942 \\
			T$^2$Net \cite{zheng2018t2net} & 0.173 & 1.396 & 6.041 & 0.251 & 0.757 & 0.916 & 0.966 \\
			Akada \textit{et al.} \cite{akada2021self} & 0.168 & 1.228 & 5.498 & 0.235 & 0.771 & 0.921 & 0.973 \\
			DESC \cite{lopez2020desc} & 0.156 & 1.067 & 5.628 & 0.237 & 0.787 & 0.924 & 0.970 \\
			GASDA \cite{zhao2019geometry} & 0.149 & 1.003 & 4.995 & 0.227 & 0.824 & 0.941 & 0.973 \\
			SharinGAN \cite{pnvr2020sharingan} & \textbf{0.116} & \textbf{0.939} & \textbf{5.068} & \textbf{0.203} & \textbf{0.850} & \textbf{0.948} & \textbf{0.978} \\			
			\noalign{\smallskip} \hline \noalign{\smallskip}
			LFDA (ours) & \textbf{0.120} & \textbf{0.961} & \textbf{5.095} & \textbf{0.213} & \textbf{0.848} & \textbf{0.945} & \textbf{0.975} \\
			\noalign{\smallskip} \hline
		\end{tabular}
	\end{center}
	\vspace*{-\baselineskip}
\end{table}

\subsection{Synthetic-to-Real Adaptation}
The style and appearance of synthetic images are usually different from that of real images. This can negatively impact the accuracy on real data.

\noindent \textbf{Datasets.}
We use Virtual KITTI (vKITTI) \cite{gaidon2016virtual} and KITTI as the source and the target domains, respectively. Following \cite{zheng2018t2net}, we resize the vKITTI images to 192×640 and cap the ground-truth depth at 80m. We evaluate on both KITTI Eigen split and KITTI stereo 2015 dataset \cite{menze2015object}.

\noindent \textbf{Results.}
Table~\ref{table:kitti2015} reports the test results on KITTI stereo 2015 dataset. We also put our Cityscapes-to-KITTI model for comparison. As it can be observed, both our models achieve much better accuracy than present approaches in most metrics. Note that Atapour \textit{et al.} \cite{atapour2018real} uses the images captures from the GTA5 game as their source data, and KITTI has a smaller domain shift with GTA5 than Cityscapes or vKITTI. Table~\ref{table:vkitti2kitti} shows the test results on KITTI Eigen split. LFDA significantly outperforms most existing works, while it is behind SharinGAN \cite{pnvr2020sharingan} by a slim margin. Note that SharinGAN requires a sophisticated image translation network during inference, resulting in a much higher computational cost than ours. Also, it relies on a complicated multi-stage training procedure. Both drawbacks make it unfriendly to be deployed in real-world applications.

\subsection{Adverse Weather Adaptation}
Adverse weather such as fog and rain produce image artifacts. These artifacts can result in accuracy degradation.

\noindent \textbf{Datasets.}
In this experiment, Foggy Cityscapes \cite{sakaridis2018semantic} is used as the target dataset. It is constructed by simulating haze upon Cityscapes images. We crop and resize the images to 192×640, and cap the ground-truth depth at 80m.

\noindent \textbf{Results.}
Table~\ref{table:foggy} reports the test results on Foggy Cityscapes. We evaluate both our models of Cityscapes-to-KITTI and vKITTI-to-KITTI. Since this is the first time in the literature to explore adverse weather adaptation for MDE, we build our own baselines to compared with. Src-Only refers to the model trained on only the source data, and Src+Tgt+AL is trained on both source and target data by adversarial learning to align their entire feature distributions. Clearly, LFDA makes considerable improvements over both baselines, indicating that it performs more stably under different weather conditions. Examples of qualitative results are shown in Fig.~\ref{fig:cover}.

\setlength{\tabcolsep}{2.5pt}
\begin{table}[t!]
	\scriptsize
	\begin{center}
		\caption{Results on Foggy Cityscapes \cite{sakaridis2018semantic} (cap 80m).}
		\label{table:foggy}
		\begin{tabular}{l | c | cccc | ccc}
			\hline \noalign{\smallskip} \noalign{\smallskip}
			&  &  & \multicolumn{2}{c}{\underline{\hspace{0.4em} Lower, better \hspace{0.4em}}} &  & \multicolumn{3}{c}{\underline{\hspace{0.4em} Higher, better \hspace{0.4em}}} \\
			Method & Dataset & abs-rel & sq-rel & rmse & rmse-log & 1.25 & 1.25$^2$ & 1.25$^3$ \\
			\noalign{\smallskip} \hline \noalign{\smallskip}
			Src-Only & CS & 0.477 & 8.333 & 18.211 & 0.717 & 0.225 & 0.507 & 0.720 \\
			Src+Tgt+AL & CS \& K & 0.422 & 4.672 & 11.879 & 0.448 & 0.249 & 0.698 & \textbf{0.915} \\
			LFDA (ours) & CS \& K & \textbf{0.283} & \textbf{3.485} & \textbf{11.26}1 & \textbf{0.381} & \textbf{0.479} & \textbf{0.835} & 0.914 \\
			\noalign{\smallskip} \hline \noalign{\smallskip}
			Src-Only & vK & 0.415 & 9.117 & 17.356 & 0.673 & 0.370 & 0.631 & 0.741 \\
			Src+Tgt+AL & vK \& K & 0.378 & 6.130 & 15.434 & 0.600 & 0.325 & 0.688 & 0.795 \\
			LFDA (ours) & vK \& K & \textbf{0.332} & \textbf{4.454} & \textbf{13.024} & \textbf{0.475} & \textbf{0.374} & \textbf{0.762} & \textbf{0.868} \\
			\noalign{\smallskip} \hline
		\end{tabular}
	\end{center}
	\vspace*{-\baselineskip}
\end{table}

\subsection{Ablation Study}
We conduct an ablation study using our model of vKITTI-to-KITTI and evaluate on the KITTI Eigen split. The results are reported in Table~\ref{table:ablation}. First, we can see that +Tgt+AL makes an obvious improvement over Src-Only, showing the importance of domain adaptation. Second, +Tgt+Con+2BN refers to the model that makes use of the decomposed content features and deploys two separate BN branches for the source and target domains, respectively. +Tgt+Con+2BN greatly improves the abs-rel metric by 0.017, showing our feature decomposition and separate BNs are effective in learning the domain-invariant content feature. Next, +Tgt+Con+2BN+Sty includes $z^t_{sty}$ in the pipeline but still maintains only two separate BNs. Results show that it suffers from severe performance degradation. This proves our argument that content and style features have different distributions, so passing them through the same BN would drop model performance. Finally, LFDA (i.e. +Tgt+Con+3BN+Sty), which deploys the third BN for the target style feature exclusively, resolves this issue successfully. Obviously, LFDA performs the best in most metrics, demonstrating the effectiveness of our method.

\setlength{\tabcolsep}{3pt}
\begin{table}[t!]
	\scriptsize
	\begin{center}
		\caption{Results of ablation study, tested vKITTI-to-KITTI adaptation on KITTI Eigen split \cite{eigen2014depth} (cap 80m).}
		\label{table:ablation}
		\begin{tabular}{l | cccc | ccc}
			\hline \noalign{\smallskip} \noalign{\smallskip}
			&  & \multicolumn{2}{c}{\underline{\hspace{0.4em} Lower, better \hspace{0.4em}}} &  & \multicolumn{3}{c}{\underline{\hspace{0.4em} Higher, better \hspace{0.4em}}} \\
			Method & abs-rel & sq-rel & rmse & rmse-log & 1.25 & 1.25$^2$ & 1.25$^3$ \\
			\noalign{\smallskip} \hline \noalign{\smallskip}
			Src-Only & 0.212 & 2.196 & 7.114 & 0.323 & 0.673 & 0.851 & 0.930 \\
			+Tgt+AL & 0.140 & 1.022 & 5.131 & 0.216 & 0.834 & 0.943 & \textbf{0.977} \\
			+Tgt+Con+2BN & 0.123 & 1.039 & 5.220 & 0.215 & 0.847 & 0.944 & 0.974 \\
			+Tgt+Con+2BN+Sty & 0.273 & 3.566 & 8.371 & 0.314 & 0.659 & 0.882 & 0.948 \\
			\noalign{\smallskip} \hline \noalign{\smallskip}
			LFDA (ours) & \textbf{0.120} & \textbf{0.961} & \textbf{5.095} & \textbf{0.213} & \textbf{0.848} & \textbf{0.945} & 0.975 \\
			\noalign{\smallskip} \hline
		\end{tabular}
	\end{center}
	\vspace*{-\baselineskip}
\end{table}

\begin{figure}
	\centering
	\includegraphics[width=0.48\textwidth]{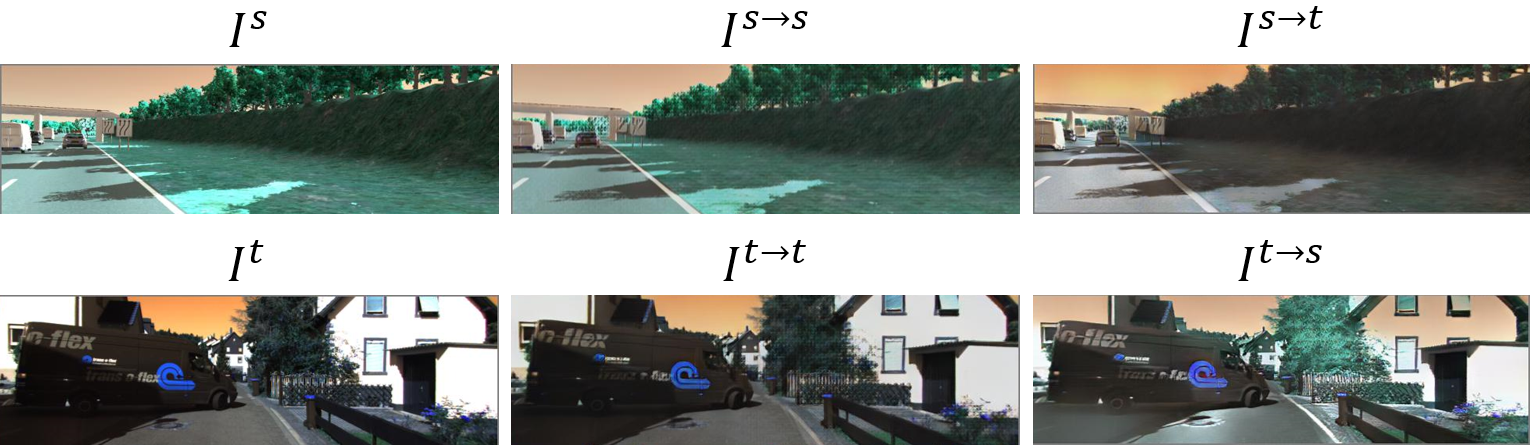}
	\vspace*{0.01\baselineskip}
	\caption{Qualitative results of the image reconstruction and translation used for feature decomposition. $I^s$: source input image (vKITTI), $I^t$: target input image (KITTI), $I^{s \ssto s}$: source reconstruction, $I^{t \ssto t}$: target reconstruction, $I^{s \ssto t}$: source-to-target translation, $I^{t \ssto s}$: target-to-source translation.}
	\label{fig:adain}
\end{figure}

\subsection{Feature decomposition visualization}
To verify the effectiveness of our feature decomposition, Fig.~\ref{fig:adain} shows the qualitative results of the image reconstruction and translation that are illustrated in Fig.~\ref{fig:big_pic}b. We can observe that the reconstructed images $I^{s \ssto s}$ and $I^{t \ssto t}$ are very close to the input images $I^s$ and $I^t$, respectively. In addition, the translated images $I^{s \ssto t}$ and $I^{t \ssto s}$ accurately maintain the content information while generating the style appearance of another domain. Such high-quality results can be achieved only if the decomposition of content and style features is successful. This demonstrates the rationale behind the high performance of the proposed method.

\subsection{Computational Complexity}
In addition to accuracy, model size and computational cost are also important factors when we evaluate a model. They determine the feasibility of a model for practical applications. In Table~\ref{table:complexity}, we compare LFDA to two existing top-performing approaches in terms of the number of parameters and the number of multiply-accumulate operations (MACs) used at inference time. GASDA \cite{zhao2019geometry} include three sub-networks during inference, a target data MDE network, a target-to-source translation network, and a target-to-source MDE network. This design places a heavy computational burden. SharinGAN \cite{pnvr2020sharingan} also needs an image translation network plus a MDE network. In contrast, in LFDA, the only sub-network in addition to the primary MDE network is $E^t_{sty}$, which increases minimum complexity. LFDA's number of MACs is 51\% and 20\% fewer than GASDA and SharinGAN, respectively, showing that our method can bridge the domain gap much more efficiently.

\setlength{\tabcolsep}{12pt}
\begin{table}[t!]
	\scriptsize
	\begin{center}
		\caption{Comparison of model complexity. The number of multiply-accumulate operations (MACs) is computed on the input size of 192×640.}
		\label{table:complexity}
		\begin{tabular}{l | cc}
			\hline \noalign{\smallskip} \noalign{\smallskip}
			Method & Params & MACs \\
			\noalign{\smallskip} \hline \noalign{\smallskip}
			GASDA \cite{zhao2019geometry} & 112.3M & 221.5G \\
			SharinGAN \cite{pnvr2020sharingan} & 57.7M & 148.1G \\
			\noalign{\smallskip} \hline \noalign{\smallskip}
			LFDA (ours) & \textbf{57.6M} & \textbf{108.1G} \\
			\noalign{\smallskip} \hline
		\end{tabular}
	\end{center}
	\vspace*{-\baselineskip}
\end{table}

\section{CONCLUSIONS}
In this paper, we propose LFDA, a novel domain adaptive MDE method. We suppose that a feature space can be decomposed into components of image content and appearance style. LFDA learns to achieve this decomposition and thus can efficiently mitigate the domain shift problem between source and target data. LFDA shows superior accuracy on three broad scenarios of domain adaptation. Moreover, it has a relatively low computational cost and can be trained end-to-end in a single stage, thereby more practical for real-world applications.



{\small
	\bibliographystyle{ieee}
	\bibliography{my_cite}
}

\end{document}